\documentclass[journal,twoside]{IEEEtran}

\usepackage{xcolor}
\usepackage{cite}
\usepackage{amsmath,amssymb,amsfonts}
\usepackage{algorithm}
\usepackage{algorithmic}
\usepackage{graphicx}
\usepackage{textcomp}
\usepackage{threeparttable}
\usepackage{array}
\usepackage{booktabs}
\usepackage{diagbox}
\usepackage{multirow}
\usepackage{subcaption}
\usepackage{makecell}
\usepackage{tabularx}
\usepackage{pifont} 

\usepackage{hyperref}
\hypersetup{hidelinks}

\graphicspath{{./figs/}}

\def\BibTeX{{\rm B\kern-.05em{\sc i\kern-.025em b}\kern-.08em
    T\kern-.1667em\lower.7ex\hbox{E}\kern-.125emX}}
\begin{document}
\title{Pixel-wise Exposure for Highly Robust In-Vehicle Remote-PPG}

\author{Jieying Wang, \IEEEmembership{Member, IEEE}, Xinqi Cai, Caifeng Shan*, \IEEEmembership{Senior Member, IEEE}, and Wenjin Wang*
	\thanks{This work is supported by the National Natural Science Foundation of China (62501366), Shandong Provincial Natural Science Foundation (ZR2024QF268), Shenzhen Key Industrial R\&D Program (ZDCYKCX20250901092803004), and Shenzhen Medical Research Fund (D2402011).}
	\thanks{This work involved human subjects in its research. Approval of all ethical and experimental procedures and protocols was granted by institutional review board of Southern University of Science and Technology under Application No. 20240150.}
	\thanks{Jieying Wang is with the College of Computer Science and Engineering, Shandong University of Science and Technology, Qingdao 266590, China (e-mail: jieying.wang@sdust.edu.cn)}
	\thanks{Wenjin Wang and Xinqi Cai are with the Department of Biomedical Engineering, College of Engineering, Southern University of Science and Technology, Shenzhen 518000, China (e-mail: wangwj3@sustech.edu.cn; 12411558@mail.sustech.edu.cn).}
	\thanks{Caifeng Shan is with the State Key Laboratory for Novel Software Technology and School of Intelligence Science and Technology, Nanjing University, Nanjing 210023, China (e-mail: cfshan@nju.edu.cn)}
	\thanks{* The corresponding authors.}
}

\maketitle

\begin{abstract}

Remote photoplethysmography (rPPG) offers a promising non-contact solution for heart rate monitoring, yet its real-world robustness is fundamentally limited by an inherent hardware limitation: existing camera exposure control paradigms—whether fixed, or auto-exposure—impose a uniform exposure time across all pixels within a frame. In high-dynamic-range scenes such as automotive cabins with strong directional sunlight, this spatially invariant exposure constraint inevitably leads to localized facial overexposure or underexposure, irreversibly corrupting the subtle pulsatile signals essential for rPPG at the point of capture—a physical degradation that no downstream algorithm can recover. 
To overcome this bottleneck, we propose PixExpo (Pixel-wise Exposure), a ``temporal-for-spatial'' framework that sequentially captures frames under a predefined cyclic exposure schedule and performs non-iterative pixel-wise fusion. 
At each pixel location, PixExpo selects the observation closest to an rPPG-motivated target intensity. This criterion seeks to reduce local saturation and severe underexposure rather than optimize perceptual appearance. PixExpo requires no sensor modification but assumes programmable frame-level exposure control.
We validate the proposed PixExpo framework using our newly introduced MEX-Drive dataset, comprising 48 participants under real-world driving conditions.
Experimental results demonstrate that PixExpo outperforms manufacture-default auto-exposure methods, reducing the mean absolute error (MAE) by 7.21\,bpm (from 13.94 to 6.73\,bpm) and increasing the success rate by 37.29 percentage points (from 25.95\% to 63.24\%) across challenging driving scenarios. 
\end{abstract}

\begin{IEEEkeywords}
Remote photoplethysmography, Multi-exposure fusion, Camera exposure control, Driver monitoring.
\end{IEEEkeywords}

\section{Introduction}
\label{sec:introduction}

\begin{figure}[!h]	
	\centerline{\includegraphics[width=0.5\textwidth]{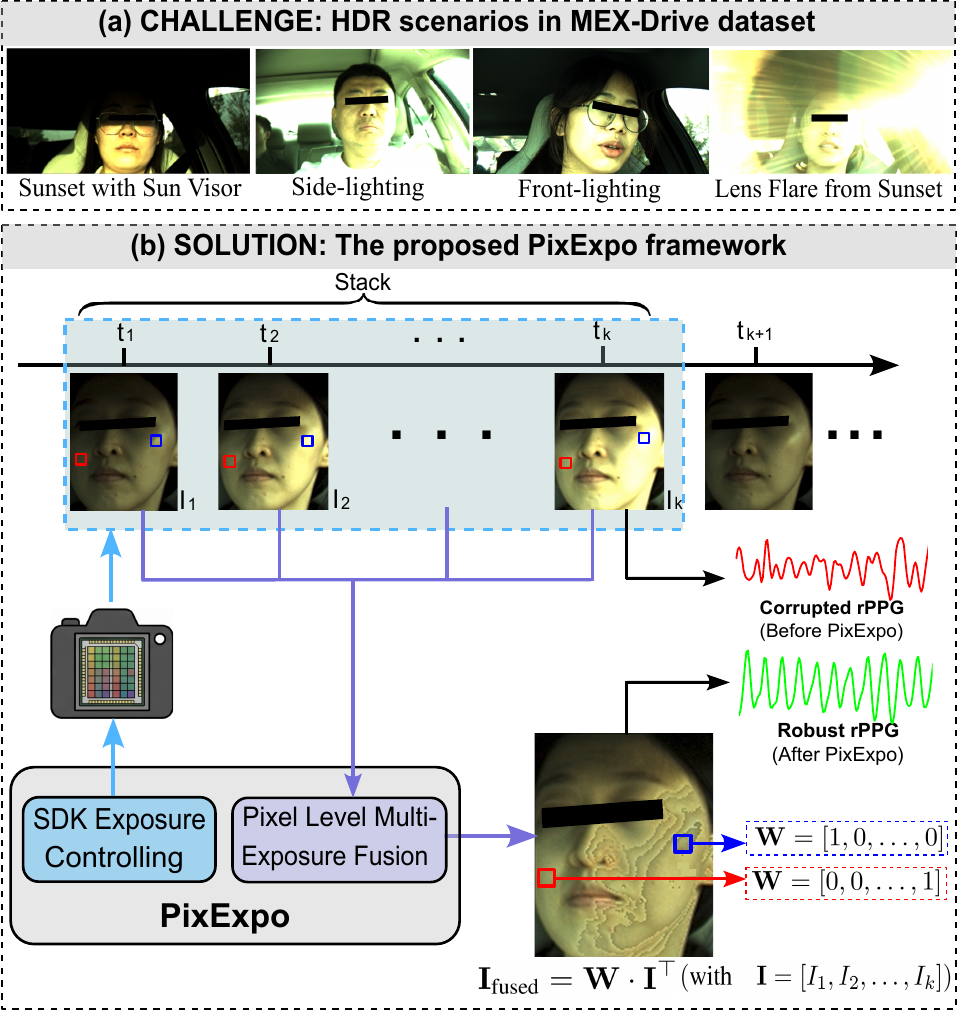}}	
	\caption{Overview of illumination challenges for in-vehicle rPPG and the proposed PixExpo framework. (a) Representative high-dynamic-range (HDR) driving scenarios from the MEX-Drive dataset. (b) Multi-exposure acquisition and pixel-wise fusion pipeline of PixExpo.}
	\label{fig_1}
\end{figure}

Remote photoplethysmography (rPPG) estimates heart rate (HR) from subtle pulse-induced skin-color changes captured by a camera, enabling contactless monitoring in telemedicine, fitness tracking, and driver monitoring~\cite{Huang2023ChallengesAP,wang2024,10856104}. However, its real-world performance remains highly sensitive to illumination~\cite{gupta2026rgb}. This challenge is particularly pronounced in vehicle cabins, where rapid lighting transitions and high-contrast shadows frequently occur, as shown in Fig.~\ref{fig_1} (a).

Critically, the vulnerability of rPPG under such conditions is not merely an algorithmic issue; it stems from a more fundamental, hardware-level limitation inherent to virtually all existing camera exposure control paradigms. Whether fixed, auto-exposure (AE), or adaptive~\cite{9126180}, every mainstream method imposes a single exposure time uniformly across all pixels within a frame.  
Under unbalanced illumination with extreme local contrast (e.g., side sunlight, shadow occlusion, and sunset glare in driving scenes), this one-exposure-fits-all mechanism makes it physically impossible to maintain optimal luminance for each skin pixel simultaneously. Inevitably, some pixels fall into overexposure (signal clipping) while others suffer from underexposure (noise dominance), resulting in irreversible corruption of the weak physiological signal. Most importantly, this imaging-source signal loss is a pure hardware/physical limitation that cannot be compensated, recovered, or repaired by any rPPG extraction or enhancement algorithms—once the photon-level information is lost due to sensor saturation, all subsequent post-processing efforts are futile. Thus, the conventional global exposure paradigm constitutes a fundamental, previously overlooked bottleneck that places a hard upper bound on the robustness of any rPPG system in real-world environments.

Existing approaches have two further limitations: (1) mainstream exposure adjustment relies on iterative fitting~\cite{kim2020proactive} or feedback control~\cite{10781984}, introducing non-negligible latency that fails to track high-frequency rapid lighting transitions in driving; (2) nearly all exposure optimization and multi-exposure fusion (MEF) criteria~\cite{ZHANG2021111,Xu2022multi,Li2021overview} are designed for human visual perception, which disrupt the subtle yet critical inter-frame chrominance variations essential for rPPG monitoring.

To address these limitations, we propose PixExpo, \textcolor{black}{a computational framework that uses a ``temporal-for-spatial" strategy to achieve pixel-wise exposure adaptation without modifying the camera's sensor architecture.} Specifically, PixExpo uses a conventional camera to rapidly capture a cyclic sequence of $K$ frames at different global exposure settings and synthesizes a fused frame by selecting the most suitable RGB observation at each pixel location, as illustrated in Fig.~\ref{fig_1}(b). \textcolor{black}{Thus, each captured frame uses a single global exposure time, while different pixels in the fused frame can originate from different exposure settings.}
This non-iterative process reduces the risk of localized saturation and severe underexposure while avoiding response delays from closed-loop exposure search. Unlike conventional MEF, PixExpo uses an rPPG-motivated photometric criterion rather than a perceptual-quality objective. 
\textcolor{black}{Notably, this framework requires no customized sensor or hardware modification, but relies on a programmable interface that supports disabling automatic exposure, reliable frame-level exposure updates, and correct association between each frame and its applied exposure setting. Such deterministic control is commonly available in industrial cameras. Consumer-grade cameras remain applicable when their vendor APIs provide these access to the full control of camera parameters programmably and a sufficiently high acquisition frame rate.}

We summarize our contributions as follows:
\begin{itemize}
	\item We identify the spatially invariant exposure constraint as a fundamental yet previously overlooked hardware-level bottleneck for rPPG robustness, and propose PixExpo, a ``temporal-for-spatial" acquisition paradigm that breaks this physical limitation via a predefined cyclic multi-exposure schedule and pixel-wise fusion, ensuring robust signal preservation in fast-varying lighting environments without iterative algorithmic latency.
	
	\item We release MEX-Drive\footnote{The dataset and source-code URLs will be made publicly available upon acceptance.}, a new dataset comprising synchronized six-level multi-exposure facial videos alongside clinical-grade ECG data from 48 participants, facilitating further research.
\end{itemize}


\section{Related work}
\label{sec2}

Remote PPG has been extensively studied for non-contact physiological monitoring, yet its deployment under challenging real-world illumination remains an open problem. While prior work has reviewed in-vehicle rPPG monitoring~\cite{10856104,10095078,s21186112,s18093080}, this section focuses specifically on exposure control strategies for robust physiological signal extraction. We organize representative methods along three dimensions: (i) spatial robustness—avoiding localized saturation or underexposure in the facial region; (ii) real-time adaptability—responding to rapid lighting changes without algorithmic latency; and (iii) rPPG-motivated design—whether the criterion targets physiological measurement rather than perceptual image quality. Table~\ref{table:comparison} provides a structured overview.

\newcommand{\cmark}{\ding{51}}
\newcommand{\xmark}{\ding{55}}

\begin{table*}[t]
	\centering
	\begin{threeparttable}
		\caption{Comparison of Existing Exposure Control Strategies and the Proposed PixExpo}
		\label{table:comparison}
		\begin{tabular}{l 
				>{\centering\arraybackslash}p{2.2cm}
				>{\centering\arraybackslash}p{3.5cm}
				>{\centering\arraybackslash}p{3.2cm}
				>{\centering\arraybackslash}p{2.8cm}}
			\toprule
			\textbf{Method Category} & \textbf{Representative Works} & \textbf{Spatial Robustness} \newline \small{(Avoids Local Saturation)} & \textbf{Real-time Adaptability} \newline \small{(Low Latency)} & \textbf{rPPG-Oriented} \\ 
			\midrule
			Standard Auto-Exposure & \cite{cho1999fast,su2016model,4682088,li2015real,shim2019gradient,mehta2020gradient,han2023camera,kim2020proactive} & \xmark \newline \small{(Global average)} & $\sim$ \newline \small{(Partially)} & \xmark \newline \small{(Vision-oriented)} \\
			\addlinespace
			Traditional MEF &~\cite{xu2022multi_exposure,Li_Kang,Li2013GuidedFiltering,Shen2011,raman2009bilateral,mertens2009exposure,GU2012604,liu2025perceptual,Fu2021auto,QU2023389,shen2022alternating} & \cmark \newline \small{(Pixel/Region-level)} & $\sim$ \newline \small{(Partially)} & \xmark \newline \small{(Vision-oriented)} \\
			\addlinespace
			rPPG-specific Global AE &~\cite{9126180,10208824,10781984,11253890,10782278,wang2026} & \xmark \newline \small{(Global average)} & $\sim$ \newline \small{(Partially)} & \cmark \newline \small{(\textbf{rPPG-oriented})} \\
			\midrule
			\textbf{Proposed framework} & \textbf{PixExpo (Ours)} & \cmark \newline \small{(\textbf{Pixel-wise fusion})} & \cmark \newline \small{(\textbf{No iterative search;} \newline \textbf{sequential capture} )} & \cmark \newline \small{(\textbf{rPPG-oriented})} \\
			\bottomrule
		\end{tabular}
		
		\begin{tablenotes}
			\small
			\item \textit{Note: \cmark, \xmark, and $\sim$ denote ``explicitly addressed", ``not explicitly addressed", and ``partially addressed", respectively.}
		\end{tablenotes}
	\end{threeparttable}
\end{table*}

\subsection{Standard Auto-Exposure Algorithms}

Conventional auto-exposure (AE) methods estimate a single exposure setting from image statistics. Early approaches used numerical root-finding methods, such as false-position and modified secant algorithms, to exploit the monotonic relationship between exposure and image brightness~\cite{cho1999fast,su2016model}. Backlight-aware methods adjust the target luminance using brightness statistics~\cite{4682088}, while hardware-oriented implementations classify exposure states by counting bright and dark pixels to reduce control latency~\cite{li2015real}.

For robotic and automotive vision, exposure objectives have been extended beyond global brightness.
\textcolor{black}{Shim et al.~\cite{shim2019gradient} introduced gradient-based feedback control for individual cameras and brightness balancing across multiple cameras.} 
Other methods combine image gradients and entropy for feature tracking~\cite{mehta2020gradient}, incorporate photometric calibration and motion-blur constraints~\cite{han2023camera}, or use Bayesian optimization to balance gradients, saturation, noise, and signal-to-noise ratio~\cite{kim2020proactive}.
These methods improve convergence, visibility, or machine-perception performance, but still produce one exposure setting per camera at each instant. Consequently, they cannot independently correct sunlit and shadowed facial regions within the same frame. Their responsiveness is also method-dependent: low-latency implementations are possible, whereas iterative optimization or feedback updates may lag abrupt illumination changes.

\subsection{Multi-Exposure Image and Video Processing}

Multi-exposure fusion (MEF) combines complementary observations captured at different exposure levels to alleviate the spatial limitations of global exposure~\cite{ZHANG2021111,Xu2022multi}. Image-based methods typically fuse an exposure bracket into a single image using spatial-domain~\cite{xu2022multi_exposure,Li_Kang,Li2013GuidedFiltering,Shen2011,raman2009bilateral}, transform-domain~\cite{mertens2009exposure,GU2012604}, or learning-based techniques~\cite{Fu2021auto,QU2023389}.

\textcolor{black}{Exposure diversity has also been explored for video enhancement. Shen et al.~\cite{shen2022alternating} combined cyclic short--long exposures with optical-flow-based alignment, restoration, and interpolation to reconstruct sharp, noise-reduced, high-frame-rate videos.} These image and video methods primarily target visual quality, which does not necessarily imply preservation of pulsation-induced temporal color variations. 

\subsection{rPPG-Specific Global Auto-Exposure}
 
Recognizing the unsuitability of vision-oriented methods, recent studies have proposed exposure control strategies explicitly optimized for rPPG signal fidelity~\cite{9126180,10208824,10781984,11253890,10782278,wang2026}. These methods incorporate physiological awareness into the control loop, for instance by maximizing the signal-to-noise ratio within the sensor's linear dynamic range~\cite{9126180}, utilizing signal quality metrics as feedback~\cite{10208824}, or employing proportional-integral-derivative (PID) controllers to stabilize facial luminance against ambient fluctuations~\cite{10781984,11253890,10782278}. A real-time triplet-frame linear fitting approach was also proposed in~\cite{wang2026} to dynamically adjust global exposure in vehicular environments.
While these approaches successfully shift the optimization target to be rPPG-motivated and partially mitigate latency through faster algorithms~\cite{wang2026}, they remain fundamentally trapped in the global exposure paradigm. They compute a single exposure time intended to optimize the average condition across the entire facial ROI. Consequently, they offer no improvement in spatial robustness—localized saturation on a sunlit cheek or underexposure on a shadowed brow remains an inevitable consequence when the intra-scene dynamic range exceeds the sensor's native capacity.

\subsection{Summary and Positioning of PixExpo}


As summarized in Table~\ref{table:comparison}, conventional AE provides global exposure control with method-dependent latency, traditional MEF handles spatial exposure variation but is primarily perception-oriented, and rPPG-specific AE remains spatially global. PixExpo combines a predefined cyclic exposure schedule with rPPG-motivated pixel-wise fusion. The predefined schedule avoids iterative exposure search, while pixel-wise selection explicitly addresses local exposure variation. Crucially, unlike traditional MEF, the fusion strategy in PixExpo is rPPG-motivated, ensuring that the physiological integrity of the skin pixels is prioritized over human visual aesthetics.

\section{METHODOLOGY}
\label{sec:3}
PixExpo is a computational pixel-wise multi-exposure framework designed for spatially nonuniform in-vehicle illumination. It approximates a spatial exposure map by sequentially acquiring globally exposed frames and selecting or weighting the observations at each pixel location.

\subsection{Theoretical Foundation}

To describe the acquisition-level limitation of global exposure control, we use a simplified image-formation model. The digital intensity at spatial location $x$ and time $t$, captured with exposure time $E$, is expressed as
\begin{equation} 
	I(x,t;E)
	=
	\mathcal{Q}\left(
	c E L(x,t)
	\left[
	R_{\mathrm{DC}}(x)+R_{\mathrm{AC}}(x,t)
	\right]
	+
	N(x,t)
	\right),
\end{equation}
where $\mathcal{Q}(\cdot)$ maps the sensor response to the processed digital intensity range, $c$ is a system response constant, $L(x,t)$ is the incident illumination, and $N(x,t)$ represents sensor and quantization noise. The skin reflectance consists of a stationary component $R_{\mathrm{DC}}(x)$ and a small pulse-induced component $R_{\mathrm{AC}}(x,t)$. In the 8-bit RGB representation used for fusion in this study, the processed intensity range is 0--255; this does not refer to the native ADC precision of the image sensor.

Within the approximately linear and unsaturated response range, the magnitude of the pulse-related intensity variation is proportional to $E L(x,t)R_{\mathrm{AC}}(x,t)$. When the pixel value is clipped at the upper limit, part or all of this temporal variation is lost. At very low exposure, the pulsatile variation may become comparable to the sensor and quantization noise, thereby reducing rPPG SNR.

Let $I_{\mathrm{DC}}(x,t;E)$ denote the local DC intensity under exposure $E$, and let $\Omega_{\mathrm{ROI}}$ denote the facial region of interest. The set of photometrically valid pixels is defined as
\begin{equation}
	\Omega_{\mathrm{valid}}(E,t)
	=
	\left\{
	x\in\Omega_{\mathrm{ROI}}
	\mid
	I_{\min}
	\le
	I_{\mathrm{DC}}(x,t;E)
	\le
	I_{\max}
	\right\},
\end{equation}
where $[I_{\min},I_{\max}]$ denotes the usable intensity range. When the effective illumination range across the face exceeds the usable dynamic range of the sensor, no single scalar exposure $E$ can place all facial locations within this range. Consequently, a global exposure setting reduces the number of skin pixels that provide photometrically suitable observations for rPPG.

An ideal spatial exposure map would select, for each location, the exposure that brings its DC intensity closest to a target operating point:
\begin{equation}
	E^{*}(x,t)
	=
	\arg\min_{E}
	\left|
	I_{\mathrm{DC}}(x,t;E)
	-
	I_{\mathrm{target}}
	\right|.
\end{equation}
Because the camera used in this study cannot assign independent exposure times to individual pixels, PixExpo approximates this ideal map through sequential global exposures and computational pixel-wise fusion.

\subsection{The Proposed PixExpo Framework}

PixExpo consists of two modules: SDK-based multi-exposure acquisition and pixel-wise multi-exposure fusion. The acquisition module repeatedly captures frames using a predefined exposure schedule, while the fusion module evaluates and combines the corresponding pixel observations.

\subsubsection{SDK-Based Multi-Exposure Acquisition}

Automatic exposure is disabled, and the camera SDK updates the exposure time on a frame-by-frame basis according to the predefined cyclic schedule
\begin{equation}
	\mathcal{E}
	=
	\{E_1,E_2,\ldots,E_N\},
	\qquad
	E_1<E_2<\cdots<E_N,
\end{equation}
where $N=6$ in the main configuration. Each acquisition cycle contains $N$ consecutive frames, with the $k$-th frame captured using exposure time $E_k$. The schedule repeats continuously without iterative exposure estimation or feedback search.

The resulting sequence samples a range of facial brightness levels within each output cycle and provides the candidate observations for subsequent fusion. Because these frames are acquired sequentially rather than simultaneously, inter-frame timing and spatial misalignment are considered in the later analysis.

\subsubsection{Pixel-Wise Multi-Exposure Fusion}

\textcolor{black}{For the $k$-th exposure frame, let $I_k(i,j)$ denote the mean RGB intensity of the pixel at the coordinate $(i,j)$ in the $k$-th sub-stream.} The deviation $D_k(x, y)$ of the $k$-th exposure from this target is calculated as:
\begin{equation}
	D_k(i,j) = | I_k(i,j) - I_{target} |.
\end{equation}
\textcolor{black}{The deviation $D_k(i,j)$ is an rPPG-motivated photometric proxy rather than a direct measure of physiological signal quality. It does not directly measure pulsatile amplitude, periodicity, chrominance variation, or rPPG SNR. Instead, it favors observations that are less likely to be affected by saturation or severe underexposure, both of which can suppress or obscure the weak pulsatile component during acquisition.}
In this study, $I_{\mathrm{target}}$ is empirically fixed at 140 on the processed 8-bit intensity scale for all recordings. This intermediate operating point provides margins against both saturation and severe underexposure.

At each pixel location, the exposure candidates are ranked according to their deviations:

\begin{equation}
	\begin{split}
		\mathbf{S}(i,j)=\{s_1,s_2,\dots,s_N\}, \quad
		\text{s.t.}\quad \\
		D_{s_1}(i,j)\le D_{s_2}(i,j)\le \dots \le D_{s_n}(i,j),
	\end{split}
\end{equation}
where $s_k(i,j)$ denotes the original exposure-channel index of the $k$-th ranked candidate. Thus, $s_1(i,j)$ identifies the exposure observation closest to $I_{\mathrm{target}}$.

The fused RGB pixel is computed as
\begin{equation}
	\mathbf{I}_{\mathrm{fused}}(i,j)
	=
	\sum_{k=1}^{N}
	w_k(i,j)
	\mathbf{I}_{s_k}(i,j),
	\qquad
	\sum_{k=1}^{N}w_k(i,j)=1.
\end{equation}

We consider three weighting strategies.

\textit{1) One-hot weighting:}
\begin{equation}
	w_1(i,j)=1,
	\qquad
	w_k(i,j)=0,\quad k=2,\ldots,N.
\end{equation}
Only the highest-ranked candidate is retained, and its complete RGB vector is copied to the fused frame.

\textit{2) Gaussian decay weighting:}
\begin{equation}
	w_k(i,j)
	=
	\frac{
		\exp\left[-(k-1)^2/(2\sigma^2)\right]
	}{
		\sum_{m=1}^{N}
		\exp\left[-(m-1)^2/(2\sigma^2)\right]
	}.
\end{equation}

\textit{3) Dynamic inverse weighting:}
\begin{equation}
	w_k(i,j)
	=
	\frac{
		\left(D_{s_k}(i,j)+\epsilon\right)^{-1}
	}{
		\sum_{m=1}^{N}
		\left(D_{s_m}(i,j)+\epsilon\right)^{-1}
	},
\end{equation}
where $\epsilon$ prevents division by zero. In the Gaussian strategy, $k$ denotes the rank after sorting by photometric deviation, rather than the original exposure-channel index.

\textcolor{black}{In this study, one-hot weighting is adopted as the default strategy. Unlike Gaussian decay and inverse weighting, which blend multiple exposure candidates to improve spatial smoothness, one-hot weighting preserves the complete RGB value of the candidate closest to the target intensity, thereby avoiding contributions from suboptimal exposures. 
This design choice is quantitatively evaluated in Sec.~\ref{v-c}.
}

\subsection{Potential Artifacts of Discrete Exposure Selection}

One-hot selection can introduce spatial discontinuities at boundaries where neighboring pixels select different exposure levels. It may also introduce temporal variation when the selected exposure channel changes across successive output frames. Static differences in regional DC intensity can be partly reduced by temporal normalization, while spatial aggregation over the facial ROI can attenuate localized discontinuities. However, temporal band-pass filtering cannot remove exposure-switching components that fall within the heart-rate frequency band.

The influence of these artifacts must therefore be evaluated empirically rather than assumed to be negligible. Secs.~V-C and V-D analyze this issue through the weighting-strategy ablation, inter-frame registration comparison, and spectral analysis of the frame switching ratio and mean selection index. PixExpo thus accepts possible visual discontinuities as a tradeoff of discrete exposure selection, while its effect on rPPG measurement is assessed using physiological metrics.

\section{Experiment Setup}

\subsection{Dataset: MEX-Drive}

\begin{figure}[!h]	
	\centerline{\includegraphics[width=0.5\textwidth]{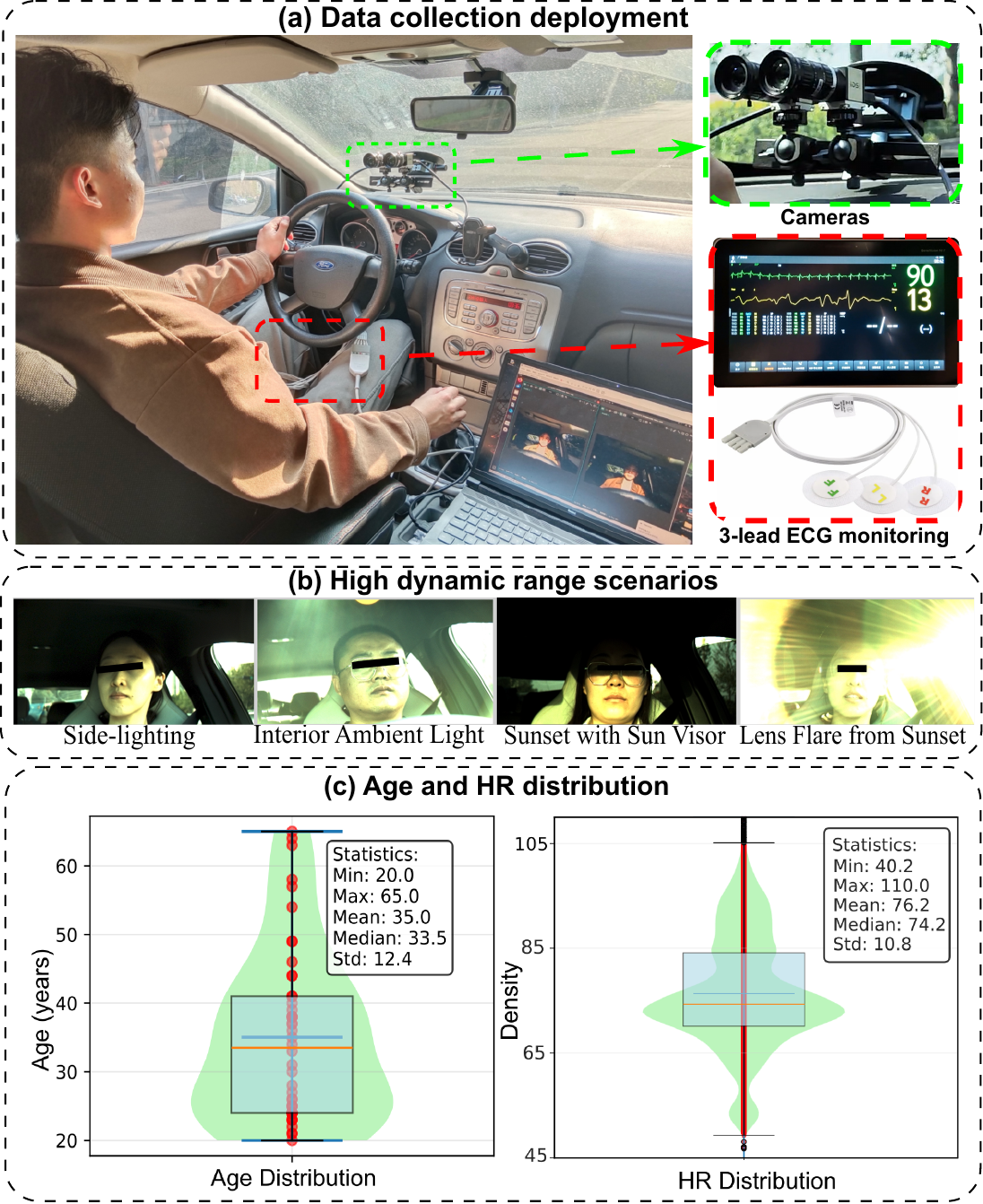}}	
	\caption{
		Experimental setup and characteristics of MEX-Drive.
		(a) Dual-camera deployment and reference ECG acquisition; 
		(b) Representative high-dynamic-range driving conditions; 
		(c) Age and reference-HR distributions of the 48 participants.
	}
	\label{fig_2}
\end{figure}

MEX-Drive was collected to evaluate multi-exposure acquisition and fusion under spatially nonuniform in-vehicle illumination. Although it used the same hardware configuration and participant pool as the ExpDrive dataset~[32], the recordings analyzed in this study were acquired using a distinct multi-exposure protocol focused on high-dynamic-range driving scenes, as shown in Fig.~\ref{fig_2}(b).

\subsubsection{Experimental Design and Protocol}

The study enrolled 48 licensed drivers, including 33 males and 15 females, aged 20--65 years ($35.0\pm12.4$ years). \textcolor{black}{
	All participants were Asian, with skin tones ranging from levels 4 to 6 on the 10-level Monk Skin Tone (MST) scale. Note that skin-tone variability was not a focus of this study, as PixExpo was designed to address dynamic illumination during driving and it does not include a mechanism specifically addressing skin-tone variation challenge in rPPG measurement.} 
	The protocol covered different illumination and weather conditions (sunny, rainy/overcast, and sunset glare), road types (highway and urban roads), three route segments (A--C), and natural driver activities such as yawning, talking, and checking the mirrors. The mean reference HR was $76.2\pm10.8$\,bpm, as shown in Fig.~\ref{fig_2}(c).
All participants received an explanation of the study and provided written informed consent. The protocol was approved by the Institutional Review Board of Southern University of Science and Technology (IRB No. 20240150).
\subsubsection{Acquisition Hardware}

\textcolor{black}{
	Two synchronized industrial RGB cameras (IDS UI-3160CP-C-HQ) were controlled through the IDS uEye API using a custom C++ acquisition program. Each $2\times2$ block of the $960\times600$ RGGB Bayer frames was converted into one RGB pixel, yielding a resolution of $480\times300$ pixels. For the cyclic multi-exposure camera, only exposure time was varied between frames; automatic exposure and automatic gain control were disabled. The sensor gain was fixed at 50, individual RGB gains were set to 0, and gain boost was disabled. Automatic white balance was disabled as well. Software gamma was set to 1.0, and hardware gamma was disabled.
}

Reference ECG signals were recorded at 500 Hz using a clinical-grade patient monitor (Mindray BeneVision N17) with a chest-lead configuration (Fig.~\ref{fig_2}(a)). The multi-exposure video sequences and ECG signals were timestamped to enable modal temporal alignment.

\subsubsection{Acquisition Configurations}

\textcolor{black}{
	All evaluated methods provided an effective video output rate of 15\,fps for rPPG signal extraction, although their acquisition rates and corresponding nominal exposure limits differed. Full-frame AE operated at 15\,fps, allowing exposures up to approximately 66.6\,ms. Triplet fitting acquired three successive frames at 45\,fps for each output frame, with a nominal exposure limit of approximately 22.2\,ms. The six-stage cyclic sequence was acquired at 90\,fps, imposing a nominal per-frame exposure limit of approximately 11.1\,ms. This limit applied to the fixed-exposure baseline, Mertens fusion, and PixExpo, all of which used recordings from this sequence.
}

\textit{Scenario I: Baseline Comparison.}
Camera 1 used the manufacturer's default full-frame auto-exposure mode, while Camera 2 acquired the six-stage cyclic exposure sequence. The fixed-exposure baseline used one pre-specified exposure substream, yielding a 15-fps video. Mertens fusion~\cite{Mertens} and PixExpo each combined all six exposure levels to produce 15-fps outputs. The input frames for Mertens fusion were spatially registered to reduce inter-frame misalignment, whereas PixExpo operated on the original, unregistered sequence in the main comparison. This scenario compared PixExpo with full-frame AE, fixed exposure, and a representative perception-oriented exposure-fusion method.

\textit{Scenario II: Comparison with Global Adaptive Exposure.}
Camera 1 implemented the triplet-frame global adaptive exposure method proposed in~\cite{wang2026}, while Camera 2 simultaneously acquired the six-stage cyclic sequence. This configuration enabled a comparison between global adaptive exposure and PixExpo under the same driving conditions.

\begin{figure*}[!t]
	\centerline{\includegraphics[width=1\textwidth]{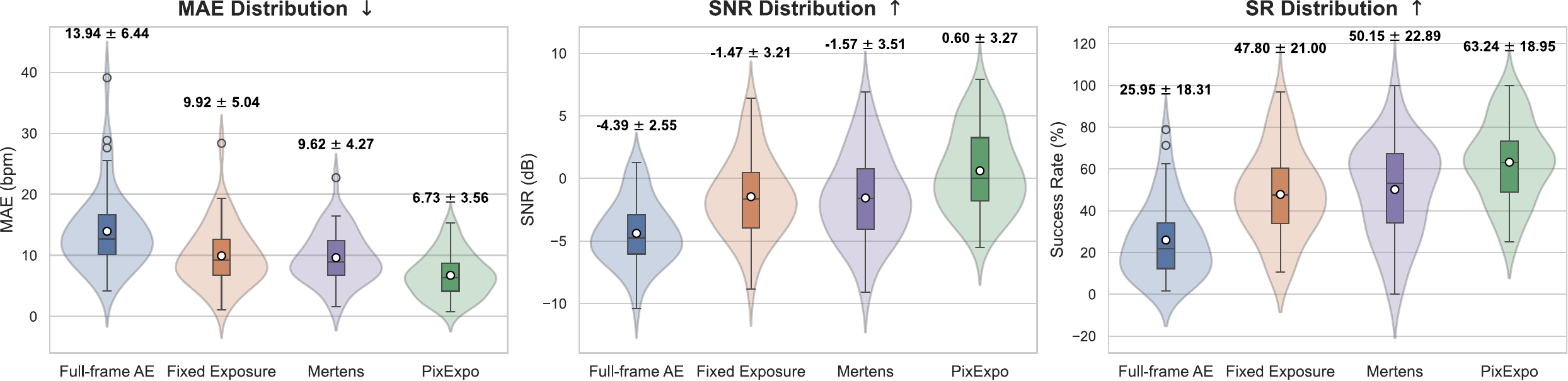}}
	\caption{\textcolor{black}{
			Participant-level distributions of MAE, SNR, and SR for four exposure and fusion strategies ($n=48$). The internal box plots show the median (red line), mean (white dot), and interquartile range.}}
	\label{fig_3}
\end{figure*}

\textcolor{black}{
\textit{Additional ROI-Based AE Experiment.}
A separate experiment with seven additional participants compared PixExpo with Face-ROI AE and Skin-ROI AE using two cameras operating concurrently. One camera acquired the six-stage PixExpo sequence, while the second camera alternated between the two ROI-based controllers, yielding a 15-fps stream for each controller. The known one-frame exposure actuation latency was explicitly accounted for in the acquisition schedule, allowing the two AE controllers to maintain independent exposure states without cross-interference.
Face-ROI AE used the YuNet face bounding box, whereas Skin-ROI AE used bilateral cheek regions localized from the five YuNet facial landmarks, thereby reducing interference from hair around the forehead. For both methods, exposure was adjusted online to maintain the mean intensity of the corresponding ROI within a predefined target range. The target range and controller parameters were fixed for all participants and were determined solely from image brightness, without using ECG, rPPG signals, or HR estimation errors for tuning. The nominal 33-ms exposure limit imposed by the 30-fps acquisition was not binding, as logged exposures for both controllers remained at or below 23 ms.
}

\subsection{Evaluation Metrics}
\textcolor{black}{
Because PixExpo targets physiological signal preservation rather than perceptual HDR quality, evaluation is based on ECG-referenced HR accuracy and rPPG signal quality. Conventional HDR image-quality metrics are not reported because they primarily assess spatial or perceptual fidelity and do not directly measure preservation of pulsation-induced temporal color variations.}
We use the metrics defined in~\cite{wang2026}:

\begin{itemize}
	\item \textit{Mean absolute error (MAE):} the mean absolute difference between the estimated HR and the ECG reference, measured in\,bpm.
	
	\item \textit{Success rate (SR):} the percentage of HR estimates satisfying
	$|H\!R_{\mathrm{est}}-H\!R_{\mathrm{ref}}|\leq5$\,bpm.
	
	\item \textit{Signal-to-noise ratio (SNR):} the ratio of spectral power within the predefined neighborhoods of the reference HR and its first harmonic to the remaining power in the $[0.7,4.0]$ Hz evaluation band.
\end{itemize}

\textcolor{black}{
POS~\cite{pos} served as the default rPPG estimator. To assess the performance consistency of PixExpo across different rPPG algorithms, EfficientPhys~\cite{liu2023efficientphys}, FactorizePhys~\cite{joshi2024factorizephys}, PhysFormer~\cite{yu2022physformer}, and iBVPNet~\cite{joshi2024ibvp} were additionally evaluated on all 48 MEX-Drive participants using PURE-pretrained checkpoints distributed by rPPG-Toolbox~\cite{liu2023rppgtoolbox}. 
Each checkpoint was applied to PixExpo, Fixed Exposure, and Full-frame AE without training or fine-tuning on MEX-Drive. The separate seven-participant ROI-based AE experiment was evaluated using POS, CHROM~\cite{dehaan2013robust}, and ICA~\cite{poh2010noncontact}.}

\textcolor{black}{
For the pretrained models, aligned RGB face crops were resized to the required spatial resolution and temporally resampled from 15 to 30\,fps by linear interpolation, preserving video duration and matching the checkpoint configurations. Model-specific normalization and temporal chunking were then applied. For each estimator, facial ROI processing, preprocessing, temporal windows, signal filtering, and HR estimation settings were kept identical across the compared acquisition strategies. 
}

\section{Results and discussion}

\begin{figure*}[!h]
	\centerline{\includegraphics[width=0.8\textwidth]{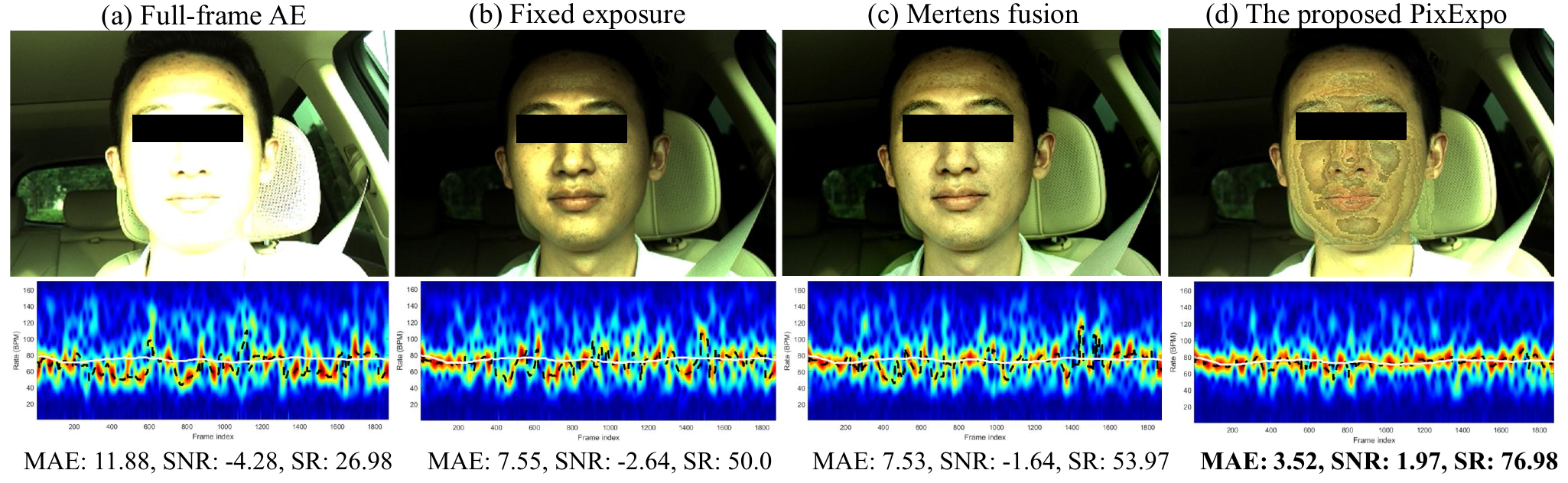}}
	\caption{\textcolor{black}{
			Visual and time-frequency comparison of four exposure and fusion strategies. The first row shows representative frames, and the second row shows the corresponding time-frequency representations. The black dashed curve denotes the estimated HR, and the white curve denotes the ECG-derived reference HR. HR MAE (bpm), SNR (dB), and SR (\%) are reported below each method.}}
	\label{fig_3-}
\end{figure*}

\subsection{Scenario I: Comparison with Baseline Strategies}

\subsubsection{Comparison with Conventional Exposure Strategies}

We first compared PixExpo with full-frame AE, the fixed-exposure substream, \textcolor{black}{and registered Mertens fusion across all 48 participants. The Mertens inputs were spatially registered before fusion to reduce the effect of inter-frame misalignment.} As shown in Fig.~\ref{fig_3}, PixExpo yielded the lowest mean MAE and the highest mean SNR and SR among the four strategies.

Specifically, PixExpo achieved an MAE of $6.73\pm3.56$\,bpm, an SNR of $0.60\pm3.27$\,dB, and an SR of $63.24\pm18.95\%$. Full-frame AE yielded $13.94\pm6.44$\,bpm, $-4.39\pm2.55$\,dB, and $25.95\pm18.31\%$, respectively. The fixed-exposure substream yielded $9.92\pm5.04$\,bpm, $-1.47\pm3.21$\,dB, and $47.80\pm21.00\%$, \textcolor{black}{while registered Mertens fusion yielded $9.62\pm4.27$\,bpm, $-1.57\pm3.51$\,dB, and $50.15\pm22.89\%$. Relative to Mertens fusion, PixExpo reduced MAE by 30.1\%, increased SNR by 2.17\,dB, and increased SR by 13.09 percentage points.} The participant-level distributions in Fig.~\ref{fig_3} are correspondingly shifted toward lower MAE and higher SNR and SR.

\textcolor{black}{
Fig.~\ref{fig_3-} provides an example. Full-frame AE produces extensive facial saturation, due to the dark cabin background driving the global controller toward a longer exposure. The fixed-exposure frame avoids severe global saturation but retains a dark forehead and bright cheeks. Registered Mertens fusion produces a more visually balanced face, whereas PixExpo exhibits visible exposure-selection boundaries.
}
\textcolor{black}{
Despite its less uniform appearance, PixExpo provides the clearest time-frequency component around the reference HR in this example. It achieves an MAE of 3.52\,bpm, an SNR of 1.97\,dB, and an SR of 76.98\%, compared with an MAE of 7.53\,bpm, an SNR of $-1.64$\,dB, and an SR of 53.97\% for Mertens fusion. Because the Mertens inputs were registered before fusion, this difference cannot be attributed solely to inter-frame misalignment. Instead, the result suggests that perceptual smoothness and preservation of pulse-related temporal variations are not equivalent objectives.}

\textcolor{black}{
Taken together, the participant-level distributions and the example show that PixExpo provides more favorable rPPG measurements under spatially and temporally varying in-vehicle illumination, although its fused frames may be less visually smooth than conventional MEF outputs.}

\textcolor{black}{
	\textit{Cross-Dataset Evaluation with Pretrained rPPG Models.}
	Table~\ref{tab:pretrained_rppg} presents results for 4 PURE-pretrained rPPG models evaluated on 48 MEX-Drive participants without fine-tuning. PixExpo achieved the lowest MAE and RMSE and the highest SR for every evaluated model. Relative to Fixed Exposure and Full-frame AE, its MAE reductions ranged from 23.1\% to 42.0\% and from 22.8\% to 53.9\%, respectively. These results show that the acquisition-related benefit of PixExpo extends to the evaluated learned estimators under cross-dataset testing.}

\begin{table}[h]
	\centering
	\caption{Cross-dataset HR estimation performance on all
		48 MEX-Drive participants using PURE-pretrained rPPG models
		without fine-tuning. Bold values indicate the best result
		among the three acquisition strategies for each estimator
		and metric.}
	\label{tab:pretrained_rppg}
	
	\setlength{\tabcolsep}{2pt}
	\renewcommand{\arraystretch}{1.0}
	\large
	
	\resizebox{0.49\textwidth}{!}{%
		\begin{tabular}{@{}l ccc ccc ccc@{}}
			\toprule
			& \multicolumn{3}{c}{PixExpo}
			& \multicolumn{3}{c}{Fixed Exposure}
			& \multicolumn{3}{c}{Full-frame AE} \\
			\cmidrule(lr){2-4}
			\cmidrule(lr){5-7}
			\cmidrule(lr){8-10}
			
			Estimator
			& MAE$\downarrow$
			& RMSE$\downarrow$ 
			& SR$\uparrow$
			& MAE$\downarrow$
			& RMSE$\downarrow$
			& SR$\uparrow$
			& MAE$\downarrow$ 
			& RMSE$\downarrow$
			& SR$\uparrow$ \\
			
			\midrule
			
			EfficientPhys~\cite{liu2023efficientphys}
			& \textbf{8.30} & \textbf{13.30} & \textbf{57.91}
			& 10.80 & 15.38 & 43.87
			& 12.05 & 16.72 & 38.21 \\
			
			FactorizePhys~\cite{joshi2024factorizephys}
			& \textbf{4.11} & \textbf{7.76} & \textbf{79.44}
			& 7.09 & 11.36 & 61.31
			& 8.91 & 13.36 & 51.04 \\
			
			PhysFormer~\cite{yu2022physformer}
			& \textbf{8.12} & \textbf{12.93} & \textbf{58.14}
			& 10.56 & 15.82 & 47.10
			& 10.52 & 15.24 & 45.27 \\
			
			iBVPNet~\cite{joshi2024ibvp}
			& \textbf{8.38} & \textbf{13.51} & \textbf{61.76}
			& 11.08 & 15.99 & 46.12
			& 12.10 & 16.71 & 40.61 \\
			
			\bottomrule
		\end{tabular}%
	}
\end{table}

\subsubsection{Additional Comparison with ROI-Based AE}
\begin{figure}[!h]
	\centerline{\includegraphics[width=0.5\textwidth]{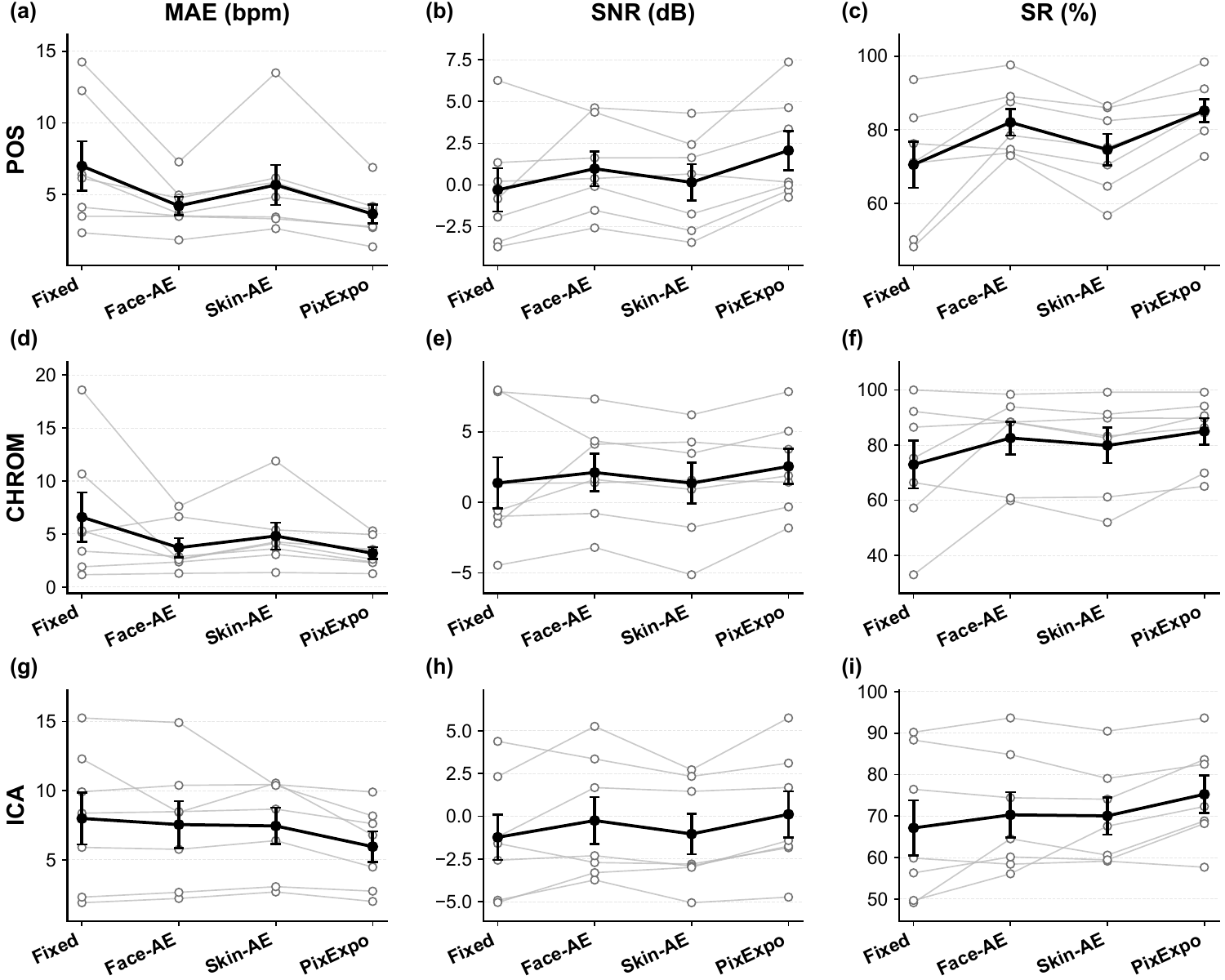}}
	\caption{\textcolor{black}{
			Performance comparison of Fixed Exposure, Face-ROI AE, Skin-ROI AE, and PixExpo using POS, CHROM, and ICA in the additional seven-participant experiment. Light lines connect paired participant-level results, and black markers show the mean $\pm$ SEM.}}
	\label{fig:roi_ae_comparison}
\end{figure}

\textcolor{black}{
Fig.~\ref{fig:roi_ae_comparison} reports the additional seven-participant comparison among the fixed-exposure baseline, Face-ROI AE, Skin-ROI AE, and PixExpo using three representative training-free rPPG algorithms: POS, CHROM, and ICA. The ROI-based AE methods generally improved upon fixed exposure, while PixExpo achieved the lowest mean MAE and the highest overall SNR and SR for all three algorithms.}

\textcolor{black}{
The mean MAEs obtained by PixExpo were 3.63, 3.19, and 5.96\,bpm for POS, CHROM, and ICA, respectively. The corresponding best ROI-based AE results were 4.21, 3.72, and 7.46\,bpm. Similar relative trends are observed for SNR and SR in Fig.~5. Within this additional experiment, these results support that the improvement provided by PixExpo is general to rPPG algorithms, not specific to POS.}

\textcolor{black}{
The remaining difference between PixExpo and ROI-based AE may be explained by the global exposure characteristics of the latter. A single exposure determined from an aggregate ROI statistic cannot independently accommodate bright and dark facial regions. Closed-loop AE may also respond to transient changes in ROI localization or illumination, causing short-term exposure fluctuations. PixExpo instead selects among multiple exposure observations at each pixel location and is therefore less dependent on a single global exposure setting.}


\subsection{Scenario II: Comparison with Fixed and Adaptive Global Exposure}

\subsubsection{Overall Performance}

\begin{figure}[!h]	
	\centerline{\includegraphics[width=0.5\textwidth]{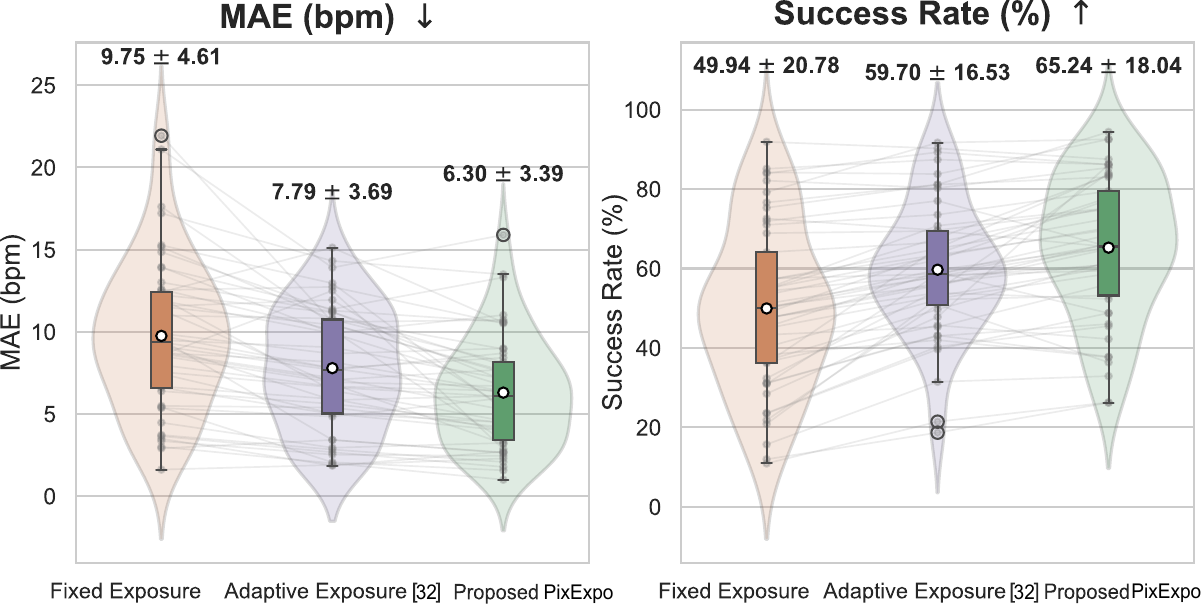}}	
	\caption{
		Participant-level comparison of fixed exposure, global adaptive exposure, and PixExpo in Scenario II ($n=48$). Grey lines connect paired results from the same participant; violin plots show the corresponding distributions of MAE and SR.
	}
	\label{fig_4}
\end{figure}

We compared PixExpo with fixed exposure and the global adaptive exposure method in~\cite{wang2026} across all 48 participants. As shown in Fig.~\ref{fig_4}, PixExpo yielded the lowest mean MAE and the highest mean SR among the three strategies.

PixExpo achieved an MAE of $6.30\pm3.39$\,bpm, compared with $7.79\pm3.69$\,bpm for global adaptive exposure and $9.75\pm4.61$\,bpm for fixed exposure. These values correspond to reductions in mean MAE of 19.1\% and 35.4\%, respectively. PixExpo also achieved an SR of $65.24\pm18.04\%$, compared with $59.70\pm16.53\%$ for adaptive exposure and $49.94\pm20.78\%$ for fixed exposure, representing increases of 5.54 and 15.30 percentage points, respectively.

The paired participant-level trajectories in Fig.~\ref{fig_4} show that PixExpo reduces MAE and increases SR for most participants relative to global adaptive exposure. However, the improvement is not uniform: several participants exhibit a higher MAE or a lower SR with PixExpo. These cases indicate that the performance of the fixed six-exposure configuration depends on the acquisition conditions. We therefore further examine the results under different illumination conditions.

\subsubsection{Performance under Different Illumination Conditions}

\begin{figure}[!h]	
	\centerline{\includegraphics[width=0.5\textwidth]{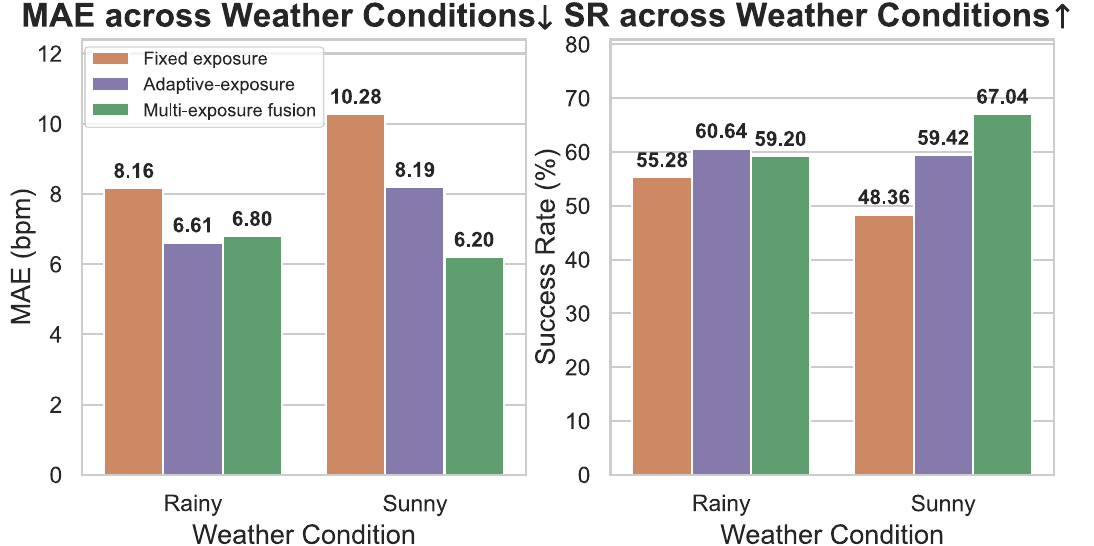}}	
	\caption{
		Mean MAE and SR of fixed exposure, global adaptive exposure, and PixExpo under rainy and sunny conditions. 
	}
	\label{fig_5}
\end{figure}

To examine how illumination affects the relative performance of the three strategies, we divided the recordings into rainy/overcast ($n=12$) and sunny ($n=36$) conditions. The subgroup means are shown in Fig.~\ref{fig_5}.

Under sunny conditions, PixExpo achieved a mean MAE of 6.20\,bpm and an SR of 67.04\%, compared with 8.19\,bpm and 59.42\% for global adaptive exposure and 10.28\,bpm and 48.36\% for fixed exposure. Direct sunlight frequently produced strong spatial illumination gradients across the face. In these conditions, the six-exposure sequence allowed PixExpo to select shorter-exposure observations for strongly illuminated regions and longer-exposure observations for darker regions, reducing localized saturation and underexposure.

Under rainy or overcast conditions, PixExpo achieved a mean MAE of 6.80\,bpm and an SR of 59.20\%. Its performance was slightly lower than that of global adaptive exposure, which achieved an MAE of 6.61\,bpm and an SR of 60.64\%, although it remained better than fixed exposure in terms of both metrics.

A plausible explanation is the exposure-time constraint imposed by the fixed six-channel configuration. Maintaining 15\,fps for each of the six exposure channels required a total acquisition rate of 90\,fps, limiting the nominal maximum exposure time to approximately 11.1 ms. The triplet-frame adaptive method operated at 45\,fps and therefore allowed a nominal maximum exposure time of 22.2 ms. This longer integration time is advantageous when photon availability is limited. These results suggest a tradeoff between spatial exposure coverage and per-frame photon accumulation, motivating an illumination-adaptive choice of exposure-channel count.

\subsubsection{Representative Case Studies}

\begin{figure}[!h]	
	\centerline{\includegraphics[width=0.5\textwidth]{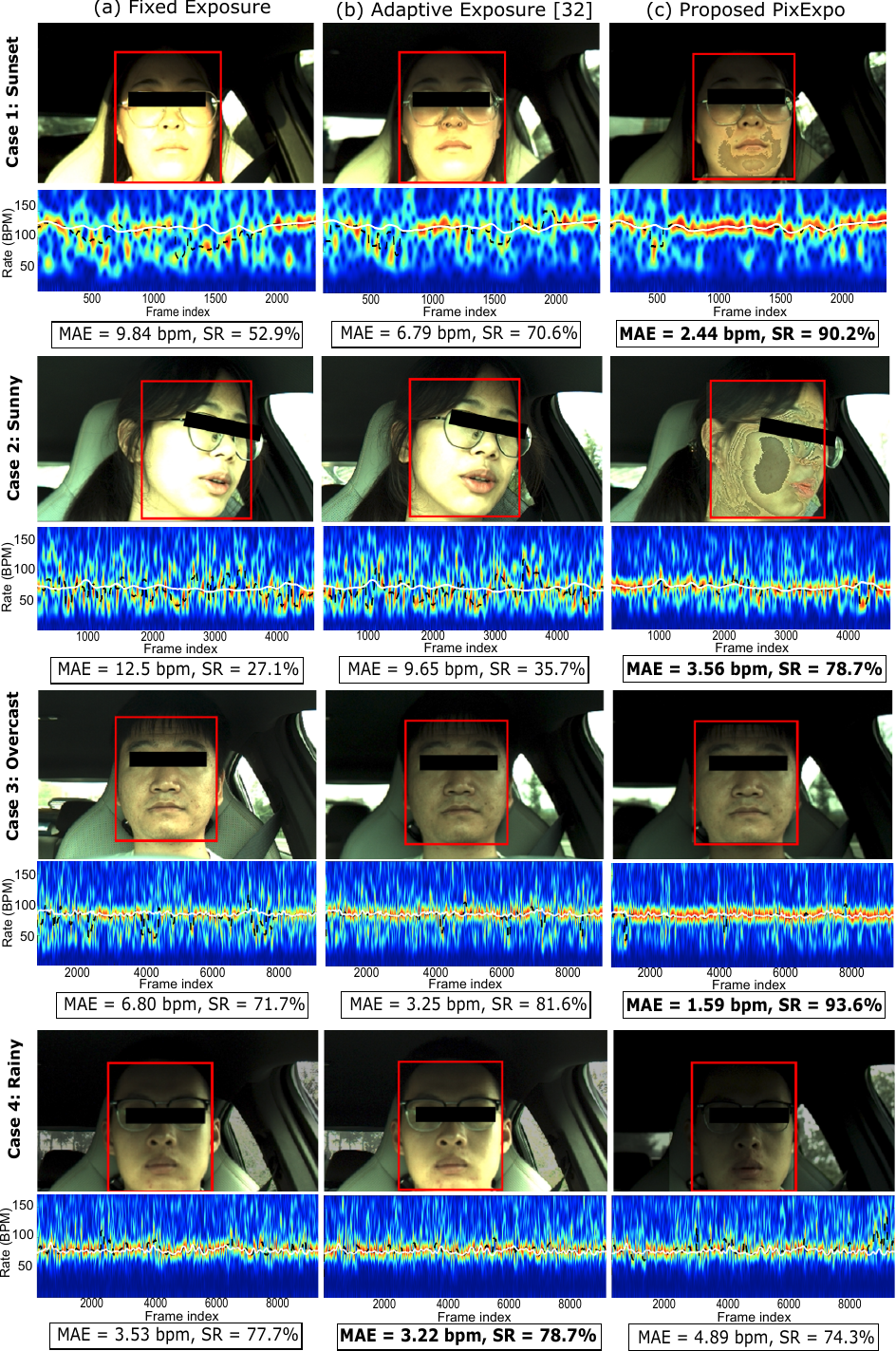}}	
	\caption{
		Representative video frames and corresponding rPPG spectrograms under four in-vehicle illumination conditions. Rows 1--4 show sunset, sunny with head rotation, overcast, and rainy conditions, respectively. Columns (a)--(c) show fixed exposure, global adaptive exposure, and PixExpo. The MAE and SR are reported below each spectrogram.
	}
	\label{fig_6}
\end{figure}

Fig.~\ref{fig_6} illustrates the behavior of fixed exposure, global adaptive exposure, and PixExpo in four representative in-vehicle conditions.

\textit{Case 1: Direct Sunset with Strong Vertical Illumination Gradients.}
Direct sunset produces a strongly illuminated lower face while the upper region remains shadowed by the vehicle roof. Fixed exposure results in extensive saturation and an unstable HR trace, with an MAE of 9.84\,bpm and an SR of 52.9\%. Global adaptive exposure moves the average facial intensity toward $I_{\mathrm{target}}=140$ and improves the result to an MAE of 6.79\,bpm and an SR of 70.6\%, but localized saturation remains around the chin and jawline. PixExpo selects shorter-exposure observations in the strongly illuminated regions, reducing local clipping and yielding an MAE of 2.44\,bpm and an SR of 90.2\%.

\textit{Case 2: Sunny Conditions with Head Rotation.}
In this case, head rotation exposes a larger cheek region to direct illumination. Fixed exposure exhibits extensive cheek saturation and achieves an MAE of 12.5\,bpm and an SR of 27.1\%. Global adaptive exposure reduces the overall exposure time, but residual local saturation remains, resulting in an MAE of 9.65\,bpm and an SR of 35.7\%. Pixel-wise one-hot selection across the six exposure levels reduces this saturation and improves the MAE to 3.56\,bpm and the SR to 78.7\%. Although the fused frame contains visible exposure-selection boundaries, these artifacts do not dominate the recovered HR trace in this example.

\textit{Case 3: Stable Overcast Conditions.}
Under relatively stable overcast illumination, all three strategies provide usable HR estimates. Fixed exposure exhibits transient deviations during changes in ambient illumination and achieves an MAE of 6.80\,bpm and an SR of 71.7\%. Global adaptive exposure produces a more continuous HR trace, with an MAE of 3.25\,bpm and an SR of 81.6\%. PixExpo yields the lowest MAE of 1.59\,bpm and the highest SR of 93.6\% in this example.

\textit{Case 4: Low-Light Rainy Conditions.}
This case illustrates a limitation of the fixed six-exposure configuration. The 90-fps acquisition rate limited the nominal maximum exposure time to 11.1 ms, reducing photon accumulation under low illumination. PixExpo achieved an MAE of 4.89\,bpm and an SR of 74.3\%, compared with 3.53\,bpm and 77.7\% for fixed exposure and 3.22\,bpm and 78.7\% for global adaptive exposure. Thus, the six-exposure configuration did not provide an advantage in this low-light example.

\subsubsection{Effect of Exposure Channel Count under Low Light}
\label{channel_count}

\textcolor{black}{
To examine whether the low-light result was related to the exposure budget, we conducted an additional controlled experiment with seven participants using $N=2$, 3, and 6 exposure channels. Each configuration was newly acquired rather than generated by subsampling the six-channel recordings. The illumination, camera settings, fusion criterion, and downstream rPPG processing were kept unchanged.}

\textcolor{black}{
As shown in Table~\ref{tab:channel_ablation}, $N=2$ achieved the lowest mean MAE of 3.75\,bpm, the highest mean SNR of 4.92\,dB, and the highest mean SR of 90.15\%. In comparison, $N=6$ yielded an MAE of 6.99\,bpm, an SNR of 1.19\,dB, and an SR of 67.76\%. Reducing the channel count from six to two decreased MAE by 46.4\%, increased SNR by 3.73\,dB, and increased SR by 22.39 percentage points.}

\begin{table}[h]
	\centering
	\caption{Effect of Exposure Channel Count under Controlled Low-Light Conditions}
	\label{tab:channel_ablation}
	\footnotesize
	\setlength{\tabcolsep}{3.5pt}
	\renewcommand{\arraystretch}{1.1}
	\begin{tabular}{cccccc}
		\hline
		$N$ &
		\shortstack{Total rate\\(fps)} &
		\shortstack{Exposure times\\(ms)} &
		\shortstack{MAE\\(bpm)} &
		\shortstack{SNR\\(dB)} &
		\shortstack{SR\\(\%)} \\
		\hline
		2 & 30 & 16, 32                         & \textbf{3.75} & \textbf{4.92} & \textbf{90.15} \\
		3 & 45 & 7, 14, 21                      & 4.59          & 3.30          & 81.76          \\
		6 & 90 & $1.8\times\{1,\ldots,6\}$      & 6.99          & 1.19          & 67.76          \\
		\hline
	\end{tabular}
\end{table}

\textcolor{black}{
These results support the explanation that a fixed six-channel configuration can become suboptimal under low illumination because the higher total frame rate restricts the available integration time. They motivate future selection of the exposure-channel count according to scene illumination. However, the seven-participant experiment does not establish that $N=2$ is universally optimal for all low-light conditions.}

\subsection{Ablation Study on Fusion Weighting Strategies}
\label{v-c}

\textcolor{black}{
To evaluate the fusion weighting strategy, we compared one-hot, Gaussian-decay, and inverse weighting in all 48 participants while keeping the multi-exposure inputs and downstream rPPG pipeline unchanged. For Gaussian-decay and inverse weighting, $\sigma=1$ and $\epsilon=1$, respectively, were fixed across participants without subject-specific tuning. As shown in Table~\ref{tab:fusion_ablation}, one-hot weighting yielded the lowest MAE ($6.73\pm3.56$\,bpm) and the highest SR ($63.24\pm18.95\%$) and SNR ($0.60\pm3.27$\,dB). Relative to Gaussian-decay weighting, it reduced MAE by 11.0\%, increased SR by 5.34 percentage points, and increased SNR by 0.45\,dB. Friedman tests showed differences among the three strategies for all metrics (all $p<0.001$), with Kendall's $W$ values of 0.365, 0.730, and 0.875 for MAE, SR, and SNR, respectively. Holm-corrected post-hoc Wilcoxon signed-rank tests further showed that one-hot weighting outperformed each soft-weighting strategy for all three metrics (all adjusted $p<0.001$).
}

\begin{table}[h]
	\centering
	\caption{Participant-level Comparison of Fusion Weighting Strategies ($n=48$)}
	\label{tab:fusion_ablation}
	\setlength{\tabcolsep}{4.5pt}
	\renewcommand{\arraystretch}{1.05}
	\begin{tabular}{lccc}
		\toprule
		\textbf{Fusion Strategy} &
		\textbf{MAE (bpm)$\downarrow$} &
		\textbf{SR (\%)$\uparrow$} &
		\textbf{SNR (dB)$\uparrow$} \\
		\midrule
		
		One-hot
		& \textbf{6.73 $\pm$ 3.56}
		& \textbf{63.24 $\pm$ 18.95}
		& \textbf{0.60 $\pm$ 3.27} \\
		
		Gaussian decay
		& 7.56 $\pm$ 3.77
		& 57.90 $\pm$ 19.25
		& 0.15 $\pm$ 3.23 \\
		
		Inverse weighting
		& 9.83 $\pm$ 4.51
		& 47.43 $\pm$ 20.73
		& $-1.34 \pm 3.52$ \\
		
		\midrule
		Friedman $p$
		& $<0.001$
		& $<0.001$
		& $<0.001$ \\
		
		Kendall's $W$
		& 0.365
		& 0.730
		& 0.875 \\
		
		\bottomrule
	\end{tabular}
	
\end{table}

\begin{figure}[h]	
	\centerline{\includegraphics[width=0.45\textwidth]{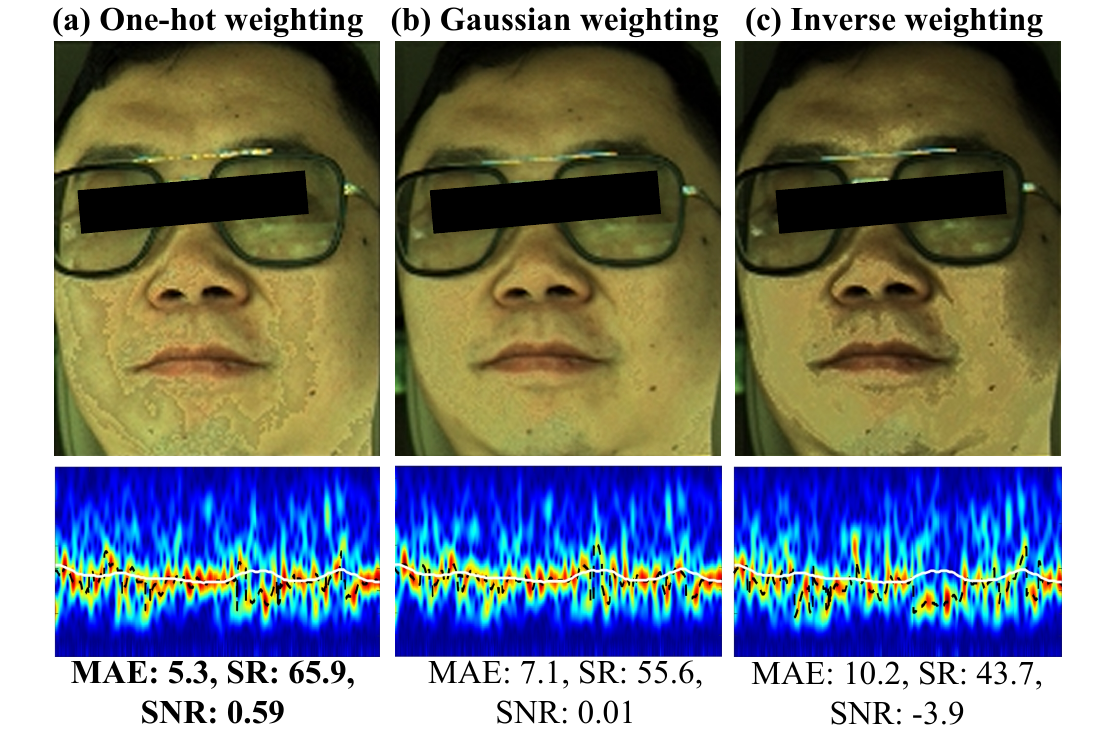}}	
	\caption{
		\textcolor{black}{Visual and time--frequency comparison of fusion weighting strategies.} 
	}
	\label{fig:fusion_ablation}
\end{figure}

\begin{figure*}[!h]
	\centerline{\includegraphics[width=1\textwidth]{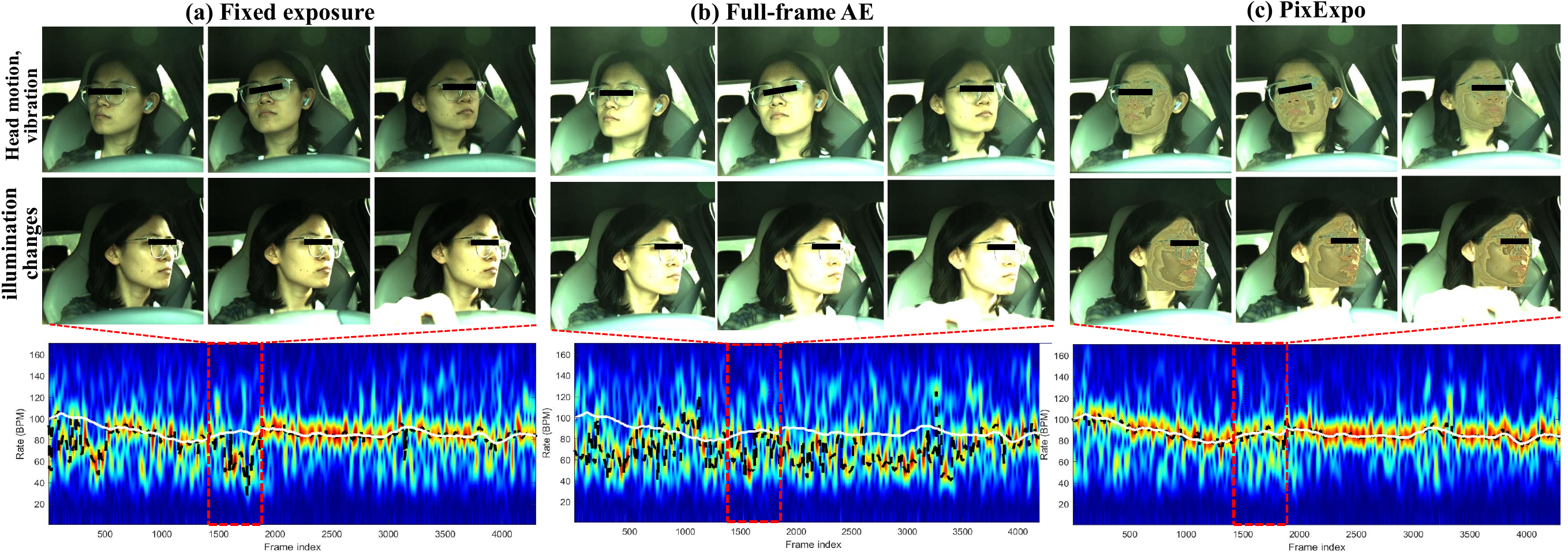}}
	\caption{\textcolor{black}{
			Representative comparison under transient driving disturbances. 
			The six frames shown above are representative frames selected from the time interval highlighted by the red dashed rectangle in each row.}}
	\label{fig_robust_all}
\end{figure*}

\textcolor{black}{
	Fig.~\ref{fig:fusion_ablation} presents a representative comparison. One-hot weighting produces sharper local intensity transitions, whereas Gaussian-decay weighting yields a smoother appearance and inverse weighting retains stronger illumination contrast. However, the time--frequency representations and physiological metrics show that greater visual smoothness does not necessarily improve rPPG performance. In the POS-based evaluation, RGB values are spatially averaged over the facial ROI before pulse extraction; smooth transitions between neighboring pixels are therefore not a prerequisite for obtaining a reliable aggregate signal. Soft weighting combines multiple temporally offset candidates, including observations ranked farther from the target intensity, which may dilute the pulsatile component. In contrast, one-hot weighting retains only the highest-ranked candidate at each pixel. These results support the use of one-hot weighting when rPPG fidelity, rather than perceptual smoothness, is the primary objective.
}

\subsection{Robustness Analysis and Limitations}

\subsubsection{Robustness to Motion and Illumination Changes}
\textcolor{black}{
	In practical driving scenarios, head motion, vehicle vibration, and rapid illumination changes may occur within one exposure cycle. Since the six exposure levels are acquired at 90 fps, one complete cycle lasts only about 66.7 ms, which limits the spatial displacement between consecutive frames under typical driving conditions. Moreover, rPPG estimation is usually based on spatially aggregated RGB signals over the skin ROI, so small local misalignments are further attenuated by spatial averaging.}
\textcolor{black}{
	Fig.~\ref{fig_robust_all} shows a representative segment containing head motion, vehicle vibration, and rapid illumination variation. Compared with fixed exposure and camera auto-exposure, PixExpo maintains a dominant spectral component close to the reference HR, demonstrating that it remains robust under these transient disturbances.}

\subsubsection{Effect of Inter-frame Registration}
\begin{figure}[!h]
	\centerline{\includegraphics[width=0.5\textwidth]{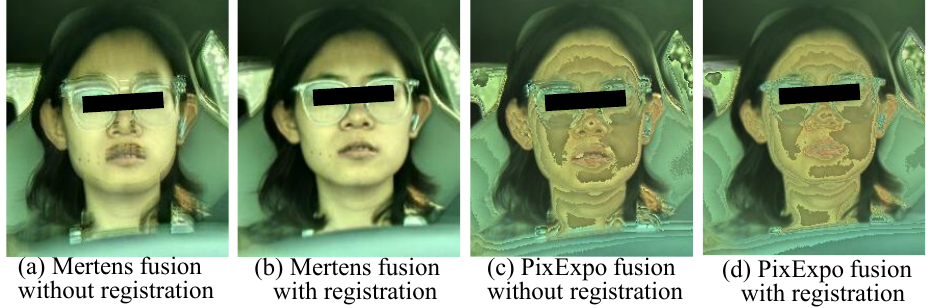}}
	\caption{\textcolor{black}{
			Visual effect of inter-frame registration on Mertens fusion and PixExpo in a representative example.}}
	\label{fig_ghost}
\end{figure}

\textcolor{black}{
Sequential multi-exposure acquisition introduces temporal offsets and potential spatial inconsistencies among the exposure candidates. Conventional multi-exposure fusion jointly blends multiple frames; therefore, misregistration can produce duplicated contours or superposition-type ghosting, as illustrated in Fig.~\ref{fig_ghost}(a). PixExpo instead selects only one exposure candidate for each output pixel. Consequently, misregistration is more likely to produce local exposure-selection discontinuities than duplicated contours.}

\textcolor{black}{
As shown in Fig.~\ref{fig_ghost}, registration visibly reduced ghosting in Mertens fusion, whereas the PixExpo outputs before and after registration remained similar. In the 14-participant subset, the mean HR MAE was $7.26\pm3.83$ bpm without registration and $7.41\pm3.83$ bpm with registration. No significant differences were found in MAE, SR, or SNR (all $p>0.05$). These results suggest that registration provided no measurable rPPG benefit under the tested conditions, although more abrupt motion warrants further evaluation.}

\subsubsection{Temporal Characteristics of Exposure Selection}
\begin{figure}[!h]
	\centerline{\includegraphics[width=0.5\textwidth]{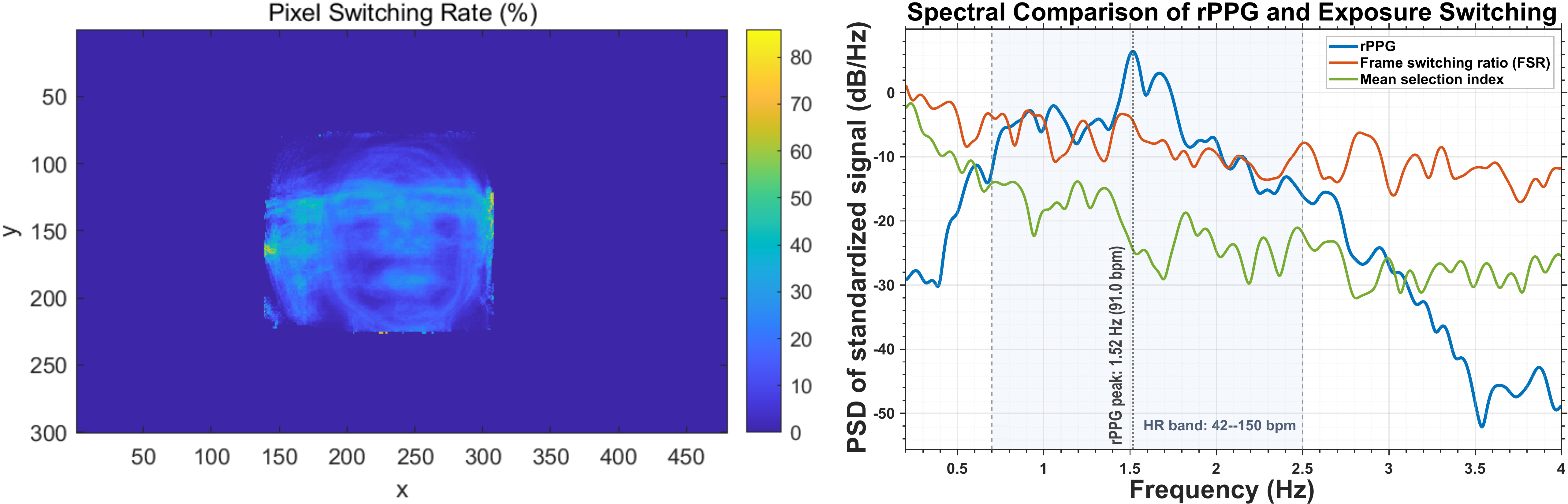}}
	\caption{\textcolor{black}{
			Spatial and spectral characteristics of exposure switching in a representative example. (a) PSR map, showing higher switching rates around facial boundaries and other high-gradient regions and lower switching rates over the forehead and cheeks. (b) PSDs of the recovered rPPG signal, FSR, and mean selection index. The recovered rPPG signal peaks at 1.52 Hz (91.0 bpm), close to the mean reference HR of 1.56 Hz.}}
	\label{fig_tt}
\end{figure}

\textcolor{black}{
Under one-hot selection, the selected exposure channel may switch over time as changes in illumination or subject movement alter the candidate intensities. Switching is particularly likely near facial boundaries and high-gradient regions, where even small spatial displacements can change pixel intensities enough to alter the selected exposure label. Consistent with this explanation, the pixel switching rate (PSR) map in Fig.~\ref{fig_tt}(a) shows more frequent switching in these regions.}

\textcolor{black}{
Frequent switching, however, does not necessarily produce periodic artifacts within the pulse-frequency range. Fig.~\ref{fig_tt}(b) compares the power spectral densities (PSDs) of the frame switching ratio (FSR), the mean selection index, and the recovered rPPG signal. FSR measures the fraction of pixels whose exposure labels change between consecutive fused frames, while the mean selection index summarizes the overall exposure-selection state. Neither switching descriptor exhibits a pronounced narrowband peak comparable to that of the recovered rPPG signal, whose peak at 1.52\,Hz is close to the mean reference HR expressed in frequency units (1.56\,Hz). These observations suggest that exposure switching is not a dominant source of pulse-like artifacts in this recording.}

\subsubsection{Limitations and Future Work}

The current implementation uses a fixed stack of six exposure levels. Although this configuration covers a broad dynamic range, the controlled low-light experiment in Sec.~\ref{channel_count} showed that it can be suboptimal under low illumination. Maintaining a 15-fps output with six exposure channels requires an aggregate acquisition rate of 90 fps, thereby limiting the maximum integration time to approximately 11.1 ms. This constraint can reduce photon accumulation and temporal SNR in dark scenes. Future work will therefore investigate illumination-aware adjustment of both the number and distribution of exposure levels.

\textcolor{black}{
	Although PixExpo targets dynamic illumination, its performance across skin tones remains to be established. The current cohort covers only MST levels 4--6, limiting the generalizability of the findings to other skin-tone groups. Future work should include participants spanning a broader range of skin tones and evaluate performance through skin-tone-stratified analyses.
}

\section{Conclusions}

This study introduced PixExpo, a computational framework that combines cyclic multi-exposure acquisition with rPPG-oriented pixel-wise fusion to mitigate spatially nonuniform illumination in vehicle cabins. Without modifying the image sensor, PixExpo selects at each pixel the exposure observation closest to a target intensity. On MEX-Drive, which included 48 participants, PixExpo reduced HR MAE from 13.94 to 6.73 bpm and increased SR from 25.95\% to 63.24\% relative to manufacturer-default auto-exposure. The additional experiments also showed better mean performance than global adaptive exposure under the tested conditions and favored one-hot selection over Gaussian-decay and inverse weighting.

The results also reveal an acquisition tradeoff. Maintaining a 15-fps output with six exposure channels requires a 90-fps acquisition rate and limits the maximum integration time to approximately 11.1 ms, which can impair performance under low illumination. Future work will adapt the number and distribution of exposure levels to scene illumination and validate PixExpo across broader skin-tone groups, additional cameras, and more severe motion conditions. Overall, PixExpo provides a practical acquisition and fusion strategy for improving in-vehicle rPPG under spatially nonuniform illumination, provided that frame-level exposure control and a sufficiently high acquisition rate are available.

\section*{References}

\bibliographystyle{elsarticle-num} 
 \bibliography{refs}

\begin{IEEEbiography}[{\includegraphics[width=1in,height=1.25in,clip,keepaspectratio]{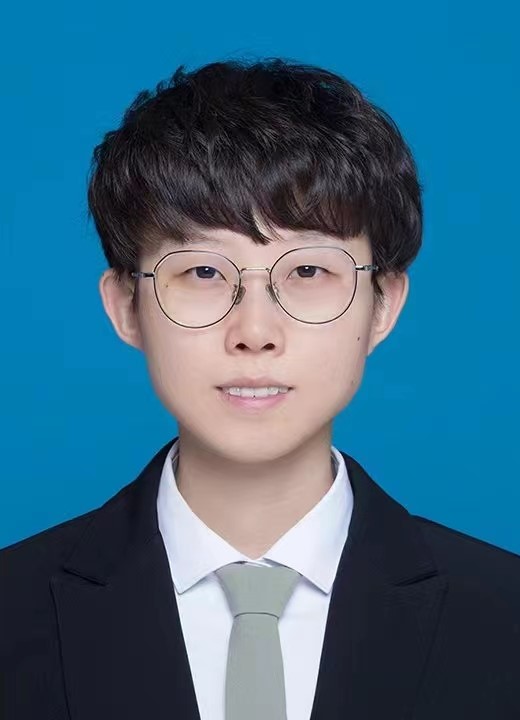}}] {Jieying Wang} received the B.E. degree in electronic and information engineering from China Jiliang University, China, in 2015, and the M.S. in marine information science and engineering from Zhejiang University, China, in 2018, and the Ph.D. degree in computer science and technology from University of Groningen, the Netherlands, in 2022. She is currently a lecture with Shandong University of Science and Technology, China. Her research interests are image processing and machine learning, with applications in health monitoring.
\end{IEEEbiography}

\begin{IEEEbiography}[{\includegraphics[width=1in,height=1.25in,clip,keepaspectratio]{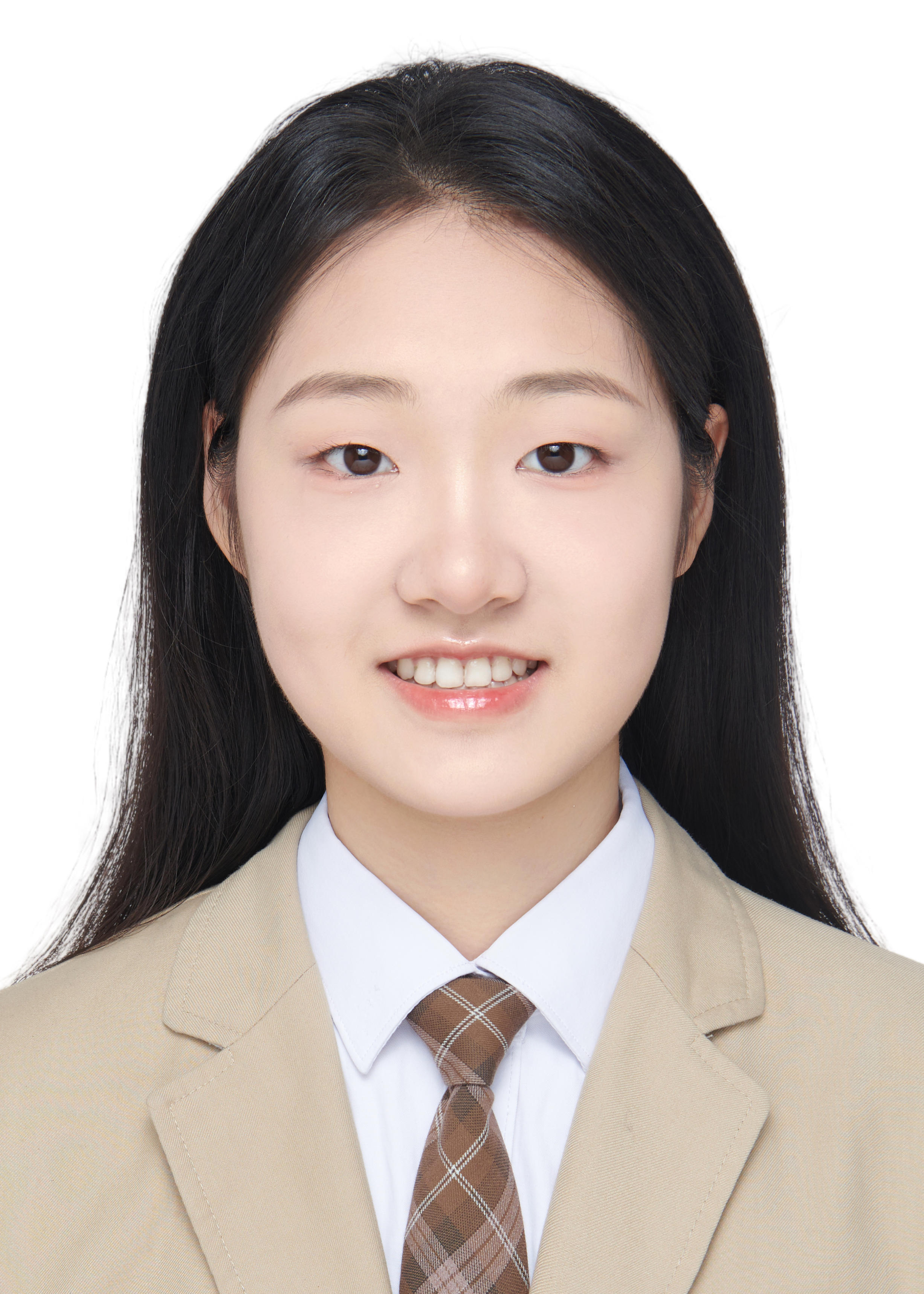}}]{Cai Xinqi} is currently an undergraduate student majoring in Biomedical Engineering at Southern University of Science and Technology (SUSTech), Shenzhen, China. Her research interests include exposure algorithms and health monitoring.
\end{IEEEbiography}

\begin{IEEEbiography}[{\includegraphics[width=1in,height=1.25in,clip,keepaspectratio]{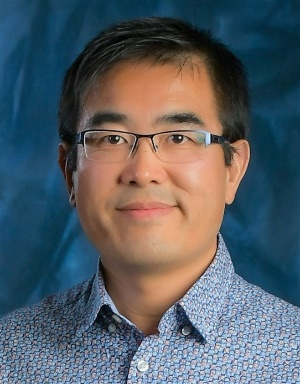}}] {Caifeng Shan} is a full professor with Nanjing University, China. He was previously a Senior Scientist with Philips Research, Eindhoven, The Netherlands. He received the B.Eng. degree from the University of Science and Technology of China, the M.Eng. degree from the Institute of Automation, Chinese Academy of Sciences, and the Ph.D. degree from Queen Mary, University of London. His research interests include computer vision, pattern recognition, medical image analysis, and related applications. He has co-authored about 200 papers and more than 100 patent applications. He has served as Associate Editor for journals including Pattern Recognition, IEEE Journal of Biomedical and Health Informatics, and IEEE Transactions on Circuits and Systems for Video Technology. He is a Senior Member of IEEE.
\end{IEEEbiography}

\begin{IEEEbiography}[{\includegraphics[width=1in,height=1.25in,clip,keepaspectratio]{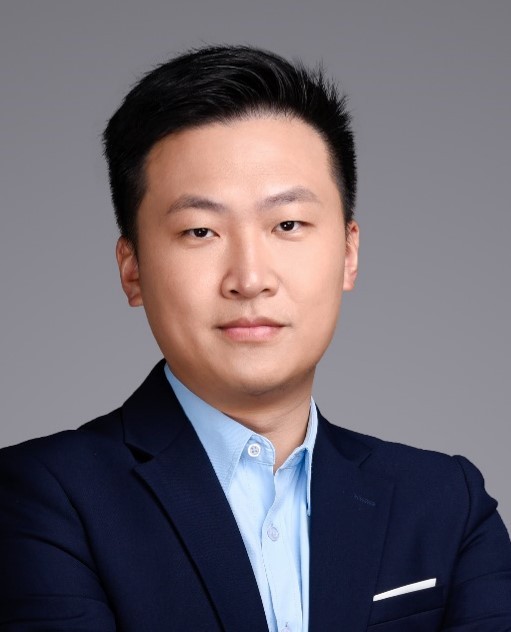}}]
	{Wenjin Wang} is an Associate Professor of Southern University of Science and Technology, China. He received the B.Sc. degree from Northeastern University, China (2011), the M.Sc. degree from University of Amsterdam, The Netherlands (2013), and the Ph.D. degree from Eindhoven University of Technology (TU/e), The Netherlands (2017). He was an Assistant Professor of TU/e and a Scientist of Philips Research Eindhoven. He published 130 peer-reviewed scientific papers, 3 academic books, and holds 45 granted patents. He got the Prize Paper Award of IEEE-TBME 2022. He was awarded the National Excellent Young Scholars (Overseas) in 2022. His current research was supported by the National Key R\&D Program of China and Natural Science Foundation of China. 
	His research interests include the video health monitoring and its translation into a medical device.
\end{IEEEbiography}

\end{document}